\documentclass{article} 
\usepackage{iclr2025_conference,times}

\usepackage{graphicx}
\usepackage{multirow}
\usepackage{booktabs}
\usepackage{adjustbox}
\usepackage{colortbl}
\usepackage{pifont}
\usepackage{soul}
\usepackage{wrapfig}
\usepackage{enumerate}
\usepackage{enumitem}

\definecolor{Blue}{rgb}{0.65, 0.70, 0.79}

\sethlcolor{Blue}

\usepackage{amsmath,amsfonts,bm}

\def\eqref#1{equation~\ref{#1}}

\def\1{\bm{1}}

\def\rvs{{\mathbf{s}}}

\def\rvv{{\mathbf{v}}}

\def\rvz{{\mathbf{z}}}

\def\rmW{{\mathbf{W}}}
\def\rmX{{\mathbf{X}}}

\DeclareMathAlphabet{\mathsfit}{\encodingdefault}{\sfdefault}{m}{sl}
\SetMathAlphabet{\mathsfit}{bold}{\encodingdefault}{\sfdefault}{bx}{n}
\newcommand{\tens}[1]{\bm{\mathsfit{#1}}}

\def\tL{{\tens{L}}}

\usepackage{hyperref}
\usepackage{url}
\hypersetup{colorlinks, breaklinks, citecolor=[rgb]{0, 0.44, 0.737},
anchorcolor=[rgb]{1, 0.44, 0.737}, linkcolor=[rgb]{0, 0.44, 0.737},
urlcolor=[rgb]{0, 0.44, 0.737} 
}

\def\equationautorefname~#1\null{Eq.~(#1)\null}
\newcommand{\aref}[1]{\hyperref[#1]{Appendix~\ref{#1}}}

\newcommand{\gptbrecq}{TesseraQ }

\makeatletter
\newif\if@restonecol
\makeatother

\usepackage[ruled,vlined]{algorithm2e}
\usepackage{algpseudocode}

\iclrfinalcopy

\title{TesseraQ: ultra low-bit llm post-training\\ quantization with block reconstruction}

\author{Yuhang Li, and Priyadarshini Panda \\
Electrical \& Computer Engineering, Yale University, USA\\
\texttt{\{yuhang.li, priya.panda\}@yale.edu}\\
\url{https://github.com/Intelligent-Computing-Lab-Yale/TesseraQ} \\
}

\begin{document}

\maketitle

\begin{abstract}
Large language models (LLMs) have revolutionized natural language processing, albeit at the cost of immense memory and computation requirements. Post-training quantization (PTQ) is becoming the \emph{de facto} method to reduce the memory footprint and improve the inference throughput of LLMs.
In this work, we aim to push the upper limit of LLM PTQ by optimizing the weight rounding parameters with the block reconstruction technique, a predominant method in previous vision models.
We propose TesseraQ, a new state-of-the-art PTQ technique, to quantize the weights of LLMs to ultra-low bits.
To effectively optimize the rounding in LLMs and stabilize the reconstruction process, we introduce progressive adaptive rounding. This approach iteratively transits the soft rounding variables to hard variables during the reconstruction process. Additionally, we optimize the dequantization scale parameters to fully leverage the block reconstruction technique.
We demonstrate that TesseraQ can be seamlessly integrated with existing scaling or clipping-based PTQ algorithms such as AWQ and OmniQuant, significantly enhancing their performance and establishing a new state-of-the-art.
For instance, when compared to AWQ, TesseraQ improves the wikitext2 perplexity from 14.65 to 6.82 and average downstream accuracy from 50.52 to 59.27 with 2-bit weight-only quantization of LLaMA-2-7B. 
{Across a range of quantization schemes, including W2A16, W3A16, W3A3, and W4A4, TesseraQ consistently exhibits superior performance.}
\end{abstract}

\vspace{-1em}
\section{Introduction}

Large Language Models (LLMs) have revolutionized natural language processing with their remarkable capabilities. LLMs such as, GPT-4~\citep{gpt4} and LLaMA-3~\citep{llama31}, contain hundreds of billions of parameters. While this scale enables their impressive performance, it also poses significant deployment challenges~\citep{zhou2024survey}. LLMs require substantial memory and computational resources, making them impractical for many real-world applications, especially on consumer devices or in resource-limited environments~\citep{llmint8}.
Quantization addresses this issue by reducing the precision of the model's parameters and activations, typically from 32-bit floating-point (FP32) to lower bit-width representations such as 8-bit or 4-bit integer (INT8, INT4). 
This technique significantly decreases the model's memory footprint to increase the I/O throughput, often with marginal performance loss. 

Post-Training Quantization (PTQ)~\citep{gholami2022survey} has perhaps become the most widespread and the easiest way to compress the LLM by reducing the bitwidth of the pretrained model's parameters. For example, with a single GPU and a small number of input sequences, GPTQ~\citep{gptq} can compress an FP16 LLM into INT4 format by deriving the exact solution for quantization error minimization. 
Recent works like AWQ~\citep{awq}, QuaRot~\citep{ashkboos2024quarot} and OmniQuant~\citep{shao2023omniquant} have pushed the compression limit further with INT3 weight-only quantization achieving a small performance gap with respect to the FP16 baseline. However, in a more challenging scenario like INT2 weight-only quantization and weight-and-activation quantization, these methods still incur a large performance gap compared to the original FP16 model. 

We conjecture that the major reason for the low performance on ultra low-bit PTQ is limited optimization space. Most works only focus on optimizing distribution transformation or weight clipping ranges~\citep{awq, outlier-plus, shao2023omniquant}. While being straightforward, they prove inadequate for extremely low-bit scenarios due to the constrained optimization space. For instance, in per-channel weight quantization, a single clipping range or transformation scale must account for 4k$\sim$20k weight elements in one channel, resulting in suboptimal quantization performance.
We contend that to enhance LLM PTQ performance further, adjustment of the entire weight tensor is necessary. However, it is non-trivial to tune billions of parameters simultaneously.

\begin{figure}[t]
\centering
\includegraphics[width=\linewidth]{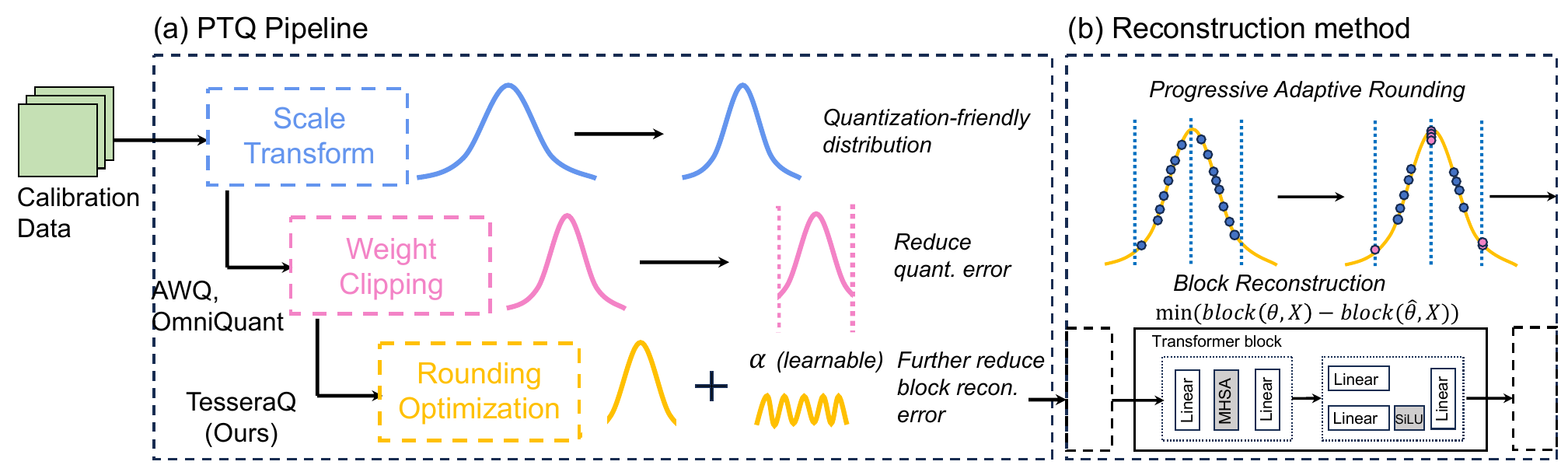}
\vspace{-2em}
\caption{\textbf{The overall workflow of our proposed method}. (a) We apply \gptbrecq to optimize the weight rounding parameters when the transformation scale and clipping range are determined using prior methods like AWQ/OmniQuant. (b) We propose Progressive Adaptive Rounding (PAR) for block-wise reconstruction, which iteratively hardens some rounding variables and optimizes the rest soft rounding variables till all variables become binary. } 
\label{fig_intro}
\vspace{-1em}
\end{figure}

To this end, we propose TesseraQ, a block reconstruction method tailored for LLM rounding optimization. We found that rounding optimization on a transformed and clipped LLM~(\autoref{fig_intro}(a)) brings significantly better performance than GPTQ. 
To accommodate the billions of parameter spaces in LLMs, our approach removes the dependency of regularization loss in the original rounding optimization processes ~\citep{adaround, brecq} by introducing Progressive Adaptive Rounding (PAR). 
As shown in \autoref{fig_intro}(b), PAR iteratively hard rounds certain rounding variables to binary values and optimizes the remainder to compensate for the rounding error. Moreover, we propose dequantization scale tuning to further decrease the reconstruction error. 
Leveraging block-wise reconstruction, we can efficiently and effectively optimize each LLM block on a single GPU. We have validated \gptbrecq across various LLMs and uniform quantization bit-widths, demonstrating superior post-training performance and establishing new state-of-the-art quantized LLMs. 
We summarize our contributions as follows
\begin{enumerate}[leftmargin=12pt]
\item We propose TesseraQ, a block reconstruction-based weight rounding optimization method for LLMs. TesseraQ can be combined with existing transformation or clipping methods like AWQ, OmniQuant, and QuaRot to obtain state-of-the-art results. 
\item TesseraQ contains Progressive Adaptive Rounding and Dequantization Scale Tuning. Both methods can stabilize the reconstruction process and effectively optimize post-training performance. 
\item Our method obtains state-of-the-art performance on both perplexity metric and zero-shot accuracy metric. For example, our method improves OmniQuant perplexity results from 37.4 to \textbf{8.0} on LLaMA-2-7B W2A16 quantization. Moreover, TesseraQ+QuaRot improves the average accuracy by \textbf{10\%} on LLaMA-3.1-8B W3A3 quantization as compared to GPTQ+QuaRot.
\end{enumerate}

\section{Preliminaries}
\label{gen_inst}

This section briefly introduces the existing research directions in LLM PTQ. We adopt uniform affine quantization, which essentially discretizes the floating-point representation of weights/activations into low-bit fixed-point representation, given by
\begin{equation}
\rmW^q = \mathrm{clamp}\biggl(\Bigl\lfloor \frac{\rmW}{s} \Bigr\rceil+z, 0, 2^N-1\biggr), \ \ \text{where }  s = \frac{\gamma \max(\rmW) - \beta \min(\rmW)}{2^N-1}, \ z = -\Bigl\lfloor \frac{\beta\min(\rmW)}{s}\Bigr\rceil.
\label{eq_quant}
\end{equation}
where $s$ and $z$ denote the quantization step size and the zero point. The resulting $\rmW^q$ is in the INT-$N$ format. To restore it back to its original range, the dequantization step is given by $\hat{\rmW} = s \times ( \rmW^q - z )$. 

\textbf{Optimization Objective. }
The plain rounding-to-nearest (RTN) method directly quantifies the model weights to integers without further optimization. However, this method usually results in significantly low task performance. To improve the LLM PTQ performance, parameters related to quantization are optimized with different objectives. For example, GPTQ~\citep{gptq} and AWQ~\citep{awq} utilize the layer-wise reconstruction objective, given by
\begin{equation}
\min_\epsilon (\tL(\theta+\epsilon) - \tL(\theta)) \approx \sum\nolimits_{\ell=1}^L \bigl|\bigl| \hat{\rmW}^{(\ell)} \rmX^{(\ell)} - \rmW^{(\ell)}\rmX^{(\ell)} \bigr|\bigr|_F^2,
\label{eq_layer_obj}
\end{equation}
where $\tL$ is the loss function parameterized by weights in the whole model $\theta$ and quantization noise $\epsilon=\hat{\theta}-\theta$. $\ell\in\{1, 2, \dots, L\}$ is the layer index and $\rmX$ is the input activations. 
While this layer-wise objective can provide efficient and exact solutions as in GPTQ, the objective does not consider inter-layer correlation like self-attention and residual connections in LLM. To this end, the block-wise reconstruction objective has been proposed~\citep{brecq}, as
\begin{equation}
\min_\epsilon (\tL(\theta+\epsilon) - \tL(\theta)) \approx \sum\nolimits_{b=1}^B \bigl|\bigl|\mathrm{block}(\hat\theta^{(b)}, \rmX^{(b)}) - \mathrm{block}(\theta^{(b)}, \rmX^{(b)})  \bigr|\bigr|_F^2.
\label{eq_block_obj}
\end{equation}
where, $\mathrm{block}$ refers to one decoder block in LLMs comprising self-attention, projection, feed-forward and normalization layers. In practice, both layer-wise and block-wise objectives enable efficient calibration on a single GPU due to their local computation attributes. However, block-wise objectives exhibit better performance than layer-wise objectives as they better approximate the global loss (i.e., \autoref{eq_block_obj} left side) by accounting for contributions from multiple layers.

\textbf{Optimization Space. }
Generally, three kinds of optimization spaces are explored in LLM PTQ, (1) the scale transformation, (2) the clipping range (i.e., finding the suitable $\gamma, \beta$), and (3) the weight values. They can be tied with either layer-wise or block-wise objectives. For instance, AWQ~\citep{awq} and OS+~\citep{outlier-plus} optimize transformation and clipping range using \autoref{eq_layer_obj}, while OmniQuant~\citep{shao2023omniquant} does similar optimization with \autoref{eq_block_obj}.
Since scale/clipping optimization methods are well-explored, in this paper, we aim to optimize weight values using block-wise objectives to further push the compression limits of LLM PTQ.

\section{TesseraQ: Ultra Low-Bit Post-Training Quantization}
\begin{wrapfigure}{r}{5.5cm}
\vspace{-1em}
\includegraphics[width=5.5cm]{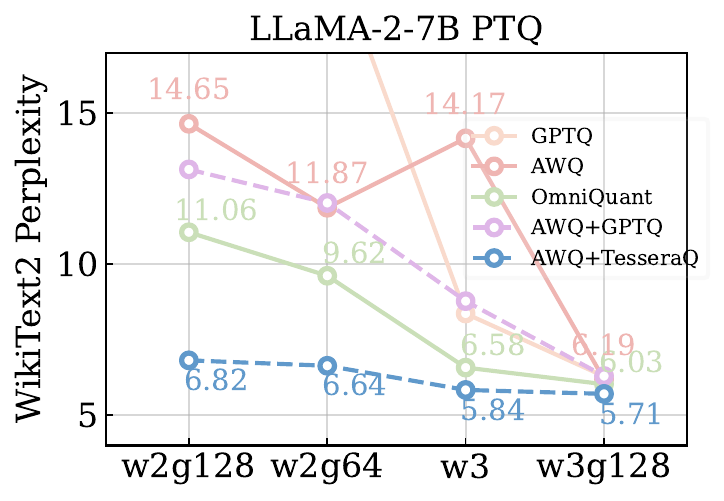}
\vspace{-1em}
\caption{Perplexity comparison of TesseraQ with other PTQ methods on LLaMA-2-7B model quantized to different weight precision (INT2, INT3). g denotes the group size.}\label{wrap-fig:1}
\vspace{-1em}
\end{wrapfigure}
\subsection{Problem Statement}

\label{sec_motivate}
Element-wise weight adjustments were also studied in GPTQ~\citep{gptq}, in which the weights are computed using closed-form solutions using the inverse Hessian matrix. 
However, we found that this technique cannot be used to improve other transformation-based PTQ methods. As shown in \autoref{wrap-fig:1} on the right side, applying GPTQ on an AWQ checkpoint barely improves the perplexity metric over AWQ across various bitwidth configurations on the LLaMA-2-7B model~\citep{llama2} 
\cite{gong2024llm} also report similar observations with GPTQ. 
We hypothesize that the reason for the failed improvement of GPTQ+AWQ could be the inaccurate layer-wise reconstruction objective and its sub-optimal solution.  
Therefore, we use weight rounding optimization~\citep{adaround, brecq}, which is a different optimization space compared to GPTQ, given by
\begin{equation}
\begin{split}
\min_{\alpha} \  & \bigl|\bigl|\mathrm{block}(\hat\theta, \rmX) - \mathrm{block}(\theta, \rmX)  \bigr|\bigr|_F^2, \\
\text{s.t.  } & 
\hat\theta = \rvs \times (\theta^q - \rvz), \ \ \theta^q = \mathrm{clamp}\Bigl(\lfloor\frac{\theta}{\rvs}\rfloor+\alpha+\rvz, 0, 2^N-1\Bigr), \ \ \alpha \in \{0, 1\}^d.
\end{split}
\end{equation}
Here, $\theta$ denotes the total $d$ weight parameters of linear layers in the block, $\alpha$ is the rounding variable. Note that we omit the block index for simplicity. This adaptive rounding optimization space restricts the range of each weight parameter. 
The restricted nature of rounding optimization will lead to more stable training and better convergence, especially when dealing with billions of parameters. 
Nonetheless, directly optimizing the rounding variable $\alpha$ is non-trivial since it is not continuous.

\subsection{Progressive Adaptive Rounding}
To optimize $\alpha$, we introduce a differentiable rounding optimization framework called progressive adaptive rounding (PAR) that does not rely on regularization loss or the straight-through estimator in contrast to previous works~\citep{adaround, adaquant}. 
To start with, we relax the rounding variable into a continuous variable by using the Sigmoid reparameterization $\alpha = \sigma(\nu)$. 
Therefore, $\nu$ can be initialized as $\nu = \sigma^{-1}(\theta / s - \lfloor\theta / s \rfloor)$, resulting in $\hat{\theta} = \theta$. 
The PAR algorithm divides all rounding variables into two sets: $\mathcal{S}_\text{Hard}$ and $\mathcal{S}_\text{Soft}$, standing for the \emph{hard} and \emph{soft} rounding of the variable $\nu$. Formally, we define the rounding function as
\begin{equation}
\alpha_i  = \begin{cases}
\sigma(\nu_i) = \frac{1}{1+\mathrm{exp}(-\nu)} & \text{if } i \in \mathcal{S}_\text{Soft} \\ 
\sigma'(\nu_i) = \mathbf{1}_{\nu_i>0} & \text{if } i \in \mathcal{S}_\text{Hard}
\end{cases}.
\end{equation}
The $\sigma'(\nu_i)$ is a hard rounding function that returns 1 if $\nu_i$ is larger than 0, otherwise it returns 0. Starting from an empty hard rounding set, we iteratively put variables  from $\mathcal{S}_\text{Soft}$ into
$\mathcal{S}_\text{Hard}$ (called \emph{Harden Phase}), and optimize the remaining soft rounding variables to compensate for the hard rounding loss (called \emph{Soften Phase}).
We elaborate on them in the following two subsections. 

\textbf{Harden Phase. }
Intuitively, after setting rounding variables to hard ones, we would expect minimum loss change in the block output error. Therefore, we define a score metric
\begin{equation}
HS(\nu) = |\sigma(\nu) - 0.5|. 
\label{eq_score}
\end{equation}
Essentially, the lower the score, the closer the soft rounding variable ($\sigma(\nu)$) is to 0.5, implying that rounding these variables to binary values will result in a larger increase in reconstruction loss. 
As a result, in the Harden Phase, we sort the parameter indices based on their $HS$ and select the lowest $P$\% of them to $\mathcal{S}_\text{Hard}$. 
The hyper-parameter $P$ should increase from 0 to near 100 during block reconstruction. During the early stage of the reconstruction, $P$ can be increased rapidly, however, in the later stage, we slowly increase $P$ since the learnable soft variables are becoming fewer in each iteration. In our experiments, we find that \gptbrecq is not sensitive to any specific decay schedule for $P$, as long as we progressively slow down the increasing rate of $P$. 
We conduct an ablation study of how to schedule the change of $P$ in \autoref{sec_ablation}. 

\begin{algorithm}[t]
    \caption{\gptbrecq Calibration process}
    \label{alg:1}
    \KwIn{FP16 LLM model; Calibration dataset, PAR iteration $K$, training steps $T$}
    \For{all $b=1,2,\dots, B$-th block in the FP model}
            {
            Collect input data to the block $\rmX$, the FP output $\mathrm{block}(\theta, \rmX)$ \;
            Initialize rounding variable $\nu$, dequantization scale $\rvv$\; 
            \For{all $k=1,2,\dots,K$-iteration}
            {   
                Calculate score (\autoref{eq_score}) and hard-round the variables with lowest $P_k$\% scores\; 
                \For{all $t=1,2,\dots, T$-training steps}{
                Gradient Descend \autoref{eq_soften} and update the soft rounding variables in this block as well as the dequantization scale\;
                }
            }
            Set all rounding variables to 0/1 and merge them into original parameters\;
            }
            \textbf{return} Quantized model\;
\end{algorithm}

\textbf{Soften Phase. }
For this stage, we employ the gradient-descent optimization to optimize the soft rounding variable
\begin{equation}
\min_{\nu_{i, i\in\mathcal{S}_\text{Soft}}} \  \bigl|\bigl|\mathrm{block}(\hat\theta, \rmX) - \mathrm{block}(\theta, \rmX)  \bigr|\bigr|_F^2.
\label{eq_soften}
\end{equation}
This objective can be optimized via gradient-based training like Adam~\citep{kingma2014adam}. 
During implementation, it would be too expensive to use masking to indicate soft rounding or hard rounding. Instead, for memory-efficient implementation, we can safely set the hard-rounding variables to $\infty$ or $-\infty$, which returns zero gradients in the sigmoid function. 
We find that optimizing \autoref{eq_soften} with nearly 200 steps can sufficiently decrease the block reconstruction error across different LLM models. 

\textbf{Post-Processing. }
After the entire PAR procedure is finished, we apply hard-rounding $\sigma'(\cdot)$ to all variables merge their values into the original weights, and then we can use the standard quantization formula (i.e., \autoref{eq_quant}). 
The merging can be effectively implemented by
\begin{equation}
\theta \leftarrow \theta + \rvs \times (\sigma'(\nu)-0.5)
\end{equation}
We provide a pseudocode for the learning process in Algorithm \ref{alg:1}.

\subsection{Dequantization Scale Tuning}

During the PAR process, the quantized tensor $\theta^q$ undergoes continuous changes. To accommodate these dynamic adjustments, we propose a method that optimizes the dequantization scale concurrently with the rounding variable. 
Specifically, for the dequantization step, we introduce an additional parameter $\rvv$ and represent it as 
\begin{equation}
\hat\theta = 2\sigma(\rvv) \times \rvs \times (\theta^q - \rvz).
\end{equation}
By initializing $\rvv=\mathbf{0}$, we initialize the dequantization scale factor ($2\sigma(\rvv)$) to 1 and subsequently adjust it to a value within the range $(0, 2)$.
Note that we avoid optimizing the scale $s$ in the quantization step (\autoref{eq_quant}) since, (1) any change in $s$ would result in a change of the rounding mechanism~\citep{adaround}, (2) the optimization requires straight-through estimation~\citep{shao2023omniquant} which leads to biased gradient calculation. 
Experiments in \autoref{sec_ablation} demonstrate that dequantization scale tuning can benefit the final quantization performance of \gptbrecq by a large margin.

\section{Experiments}

\subsection{Experiements Setup}

Most of our experiment setups are similar to OmniQuant~\citep{shao2023omniquant}, which also adopts block reconstruction loss function. Specifically, we employ asymmetric uniform quantization with 2/3/4-bit integers. We test both per-group and per-channel weight quantization. For example, we use the notation \emph{W2A16g64} to denote the 2-bit per-group (group size is set to 64) weight-only quantization, where activations are FP16. 
In weight-activation quantization experiments (all INT precision), defaults are W4A4, W3A3, and W4A8 with per-channel weight and per-token activation quantization ~\citep{llmint8, shao2023omniquant}.

\textbf{Calibration Data and Comparison. }
We report two types of evaluation metrics, the perplexity metric for evaluating the upstream datasets like WikiText2~\citep{wikitext2}, C4~\citep{c4}, and the average accuracy of 5 downstream reasoning tasks including PIQA~\citep{piqa}, ARC easy/challenge~\citep{arc}, WinoGrande~\citep{sakaguchi2021winogrande} and HellaSwag~\citep{zellers2019hellaswag}. The perplexity is evaluated with 2048 sequences.
We use 512 2048-token segments from the WikiText2 training dataset as calibration data for perplexity comparison and for downstream task comparison, we sample same amount of calibration data from the C4 training dataset. 
We use \texttt{lm\_eval} (ver0.4.2) to evaluate the accuracy.

\textbf{Training. }
We set the total PAR number of iterations $K$ to 20 and gradually increase the $P_k$ from 0 to 100\%.
In each iteration, we optimize the learnable parameters ($\nu$ and $\rvv$) for 250 training steps. We use the Adam optimizer with a fixed learning rate of $1e-3$. We add $1e-4$ weight decay to $\rvv$ during training. The batch size is set to 4. 
We use AWQ transformation~\citep{awq} to initialize our model since we find AWQ initialization is slightly better than OmniQuant across all configurations except W2A16 quantization. For W2A16, AWQ yields very high perplexity. Thus, in the W2A16 case, we directly use the pretrained OmniQuant model for initialization. 

\textbf{Models and Baselines. } For the upstream tasks, we follow OmniQuant~\citep{shao2023omniquant} to test weight-only quantization results on LLaMA-1-7B/13B/30B/65B~\citep{llama}, LLaMa-2-7B/13B/70B~\citep{llama2}. 
In this case, we compare GPTQ~\citep{gptq}, OmniQuant~\citep{shao2023omniquant}, AWQ~\citep{awq} with asymmetric clipping implementation~\citep{gong2024llm}. 
For downstream tasks, we test LLaMA-2-7B, LLaMA-3.1-8B/70B across 5 downstream tasks.
We compare GPTQ, AWQ, OmniQuant, and a recent rounding optimization method SignRound \citep{cheng2023optimize}. 

\begin{table}[t]
\setlength{\tabcolsep}{9pt}
\small
\centering
\caption{\textbf{Weight-only quantization results of LLaMA-1 and LLaMA-2 models}. We report WikiText2 perplexity (PPL $\downarrow$). *, $\dag$ means initialized from AWQ, and OmniQuant, respectively. }
\begin{adjustbox}{max width=\linewidth}
\begin{tabular}{llccccccc}
\toprule
{\textbf{LLaMA1\&2}} & \textbf{Method} & 1$-$7B & 1$-$13B & 1$-$30B & 1$-$65B & 2$-$7B & 2$-$13B &2$-$70B \\  \midrule
FP16 & - & 5.68 & 5.09 & 4.10 & 3.53 & 5.47 & 4.88 & 3.31 \\ 
\midrule
\multirow{4}{*}{\shortstack{W2A16}} 
& GPTQ & 2.1e3   &  5.5e3 & 499.75 & 55.91 & 7.7e3 & 2.1e3 & 77.95   \\
& AWQ & 1.1e5 & 7002 & 1.2e5 & 6.3e6 & 2.9e6 & 6.2e3 & 3973 \\ 
& OmniQuant & 15.47 & 13.21 & 8.71 & 7.58 & 37.37 & 17.21 & 7.81\\ 
& \cellcolor{Blue}\textbf{TesseraQ$^\dag$}  & \cellcolor{Blue}\textbf{7.56} & \cellcolor{Blue}\textbf{6.56} & \cellcolor{Blue}\textbf{5.75} &  \cellcolor{Blue}\textbf{5.21} & \cellcolor{Blue}\textbf{8.05} & \cellcolor{Blue}\textbf{6.55} & \cellcolor{Blue}\textbf{5.26}   \\
\midrule
\multirow{4}{*}{\shortstack{W2A16\\g128}} 
& GPTQ &  44.01 & 15.60 & 10.92 & 9.51 & 36.77 & 28.14 & NAN  \\
& AWQ & 13.08  & 10.02 & 7.46 & 6.08 & 14.65 & 8.93 & 5.72  \\
& OmniQuant & 9.72 & 7.93 & 7.12 & 5.95 & 11.06 & 8.26 & 6.55 \\
& \cellcolor{Blue}\textbf{TesseraQ*}  & \cellcolor{Blue}\textbf{6.92} & \cellcolor{Blue}\textbf{6.07} & \cellcolor{Blue}\textbf{5.26} & \cellcolor{Blue}\textbf{4.83} & \cellcolor{Blue}\textbf{6.82} & \cellcolor{Blue}\textbf{5.92} & \cellcolor{Blue}\textbf{4.73}  \\
\midrule
\multirow{4}{*}{\shortstack{W2A16\\g64}} 
& GPTQ & 22.10 & 10.06 &  8.54 & 8.31 & 20.85 & 22.44 & NAN \\
& AWQ &  10.65 & 8.66 & 6.65 & 5.58 & 11.87 & 7.81 & 5.30 \\
& OmniQuant & 8.90 & 7.34 & 6.59 & 5.65 & 9.62 & 7.56 & 6.11 \\
& \cellcolor{Blue}\textbf{TesseraQ*}  & \cellcolor{Blue}\textbf{6.78}  & \cellcolor{Blue}\textbf{5.97} & \cellcolor{Blue}\textbf{5.18} & \cellcolor{Blue}\textbf{4.70}   & \cellcolor{Blue}\textbf{6.67} & \cellcolor{Blue}\textbf{5.81} & \cellcolor{Blue}\textbf{4.60} \\ \midrule
\multirow{4}{*}{\shortstack{W3A16}} 
& GPTQ &  8.06 & 6.76 & 5.84 & 5.06 & 8.37 & 6.44 & 4.82 \\
& AWQ & 8.49 & 6.38 & 5.89 & 6.03 & 14.17 & 6.42 & 4.22 \\
& OmniQuant & 6.49 & 5.68 & 4.74 & 4.04 & 6.58 & 5.58 & 3.92 \\
& \cellcolor{Blue}\textbf{TesseraQ*}  & \cellcolor{Blue}\textbf{5.99} & \cellcolor{Blue}\textbf{5.35} & \cellcolor{Blue}\textbf{4.44} & \cellcolor{Blue}\textbf{3.89}  & \cellcolor{Blue}\textbf{5.84} & \cellcolor{Blue}\textbf{5.16} & \cellcolor{Blue}\textbf{3.68} \\ \midrule
\multirow{4}{*}{\shortstack{W3A16\\g128}} 
& GPTQ &  6.55 & 5.62 & 4.80 & 4.17 & 6.29 & 5.42 & 3.85 \\
& AWQ & 6.38 & 5.52 & 4.59 & 3.92 & 6.19 & 5.30 & 3.72  \\
& OmniQuant & 6.15 & 5.44 & 4.56 & 3.94 & 6.03 & 5.28 & 3.78 \\
& \cellcolor{Blue}\textbf{TesseraQ*}  & \cellcolor{Blue}\textbf{5.95} & \cellcolor{Blue}\textbf{5.32} & \cellcolor{Blue}\textbf{4.40} & \cellcolor{Blue}\textbf{3.82} & \cellcolor{Blue}\textbf{5.71} & \cellcolor{Blue}\textbf{5.11} & \cellcolor{Blue}\textbf{3.61} \\ \midrule
 \multirow{4}{*}{\shortstack{W4A16}} 
 & GPTQ & 6.13 & 5.40 & 4.48 & 3.83 & 5.83 & 5.13 & 3.58 \\
 & AWQ & 5.99 & 5.24 & 4.30 & 3.71 & 5.82 & 5.07 & 3.49 \\
 & OmniQuant & 5.86 & 5.21 & 4.25 & 3.71 & 5.74 & 5.02 & 3.47 \\
& \cellcolor{Blue}\textbf{TesseraQ*} & \cellcolor{Blue}\textbf{5.78} & \cellcolor{Blue}\textbf{5.17} & \cellcolor{Blue}\textbf{4.20} & \cellcolor{Blue}\textbf{3.63} & \cellcolor{Blue}\textbf{5.56}  & \cellcolor{Blue}\textbf{4.96} & \cellcolor{Blue}\textbf{3.40} \\ 
\bottomrule
\end{tabular}
\end{adjustbox}
\label{tab_:llama_weight_only}
\vspace{-2em}
\end{table}

\subsection{Main Results}

\textbf{Perplexity Evaluation. } 
We summarized the Wikitext2 perplexity (PPL) results in \autoref{tab_:llama_weight_only}. Our method consistently outperforms existing methods like AWQ and OmniQuant, particularly for the low-bit W2A16 configuration. Remarkably, in the W2A16 case, all existing methods except OmniQuant failed to successfully quantize the models (yielding $>1e3$ perplexity). On the LLaMA-2-7B model, OmniQuant only obtains 37.37 PPL while our method largely improves this result to \textbf{8.05}. 
We observe that in general, the lower the bitwidth, the more improvement we can obtain from \gptbrecq. This confirms our initial intuition that extremely low-bit weight quantization requires a thorough adjustment of each weight element. Additionally, the C4~\citep{c4} PPL results are provided in Appendix: \autoref{tab_:llama_weight_only_c4}. Note that the C4 results for OmniQuant are re-evaluated from the official checkpoint to align the evaluation protocol. Overall, C4 PPL results concur with the Wikitext2 results, demonstrating a similar trend in performance improvement. For example, \gptbrecq improves the PPL of LLaMA-2-7B model from 90.64 to 14.82 with W2A16 quantization.

\begin{table}[t]
\setlength{\tabcolsep}{8pt}
\small
\centering
\caption{\textbf{Weight-only quantization Results of various LLMs}. We report the accuracy of 5 common sense reasoning tasks $(\uparrow)$. * means initialized from AWQ.}
\begin{adjustbox}{max width=\linewidth}
\begin{tabular}{lclcccccc}
    \toprule
    \textbf{Models} & \textbf{Bitwidths} & \textbf{Methods} & \textbf{PiQA} & \textbf{ArcE} & \textbf{ArcC} & \textbf{HellaSwag} & \textbf{WinoGrande} & \textbf{Avg.} \\ 
    \midrule
    \multirow{11}{*}{\shortstack{LLaMA-2-7B}} &  FP16 & - & 78.07 & 76.34 & 43.51 & 57.17 & 69.21 & 64.87 \\
    \cmidrule{2-9}
    & \multirow{5}{*}{\shortstack{W2A16\\g128}} & GPTQ & 58.21 & 33.75 & 19.79 & 29.60 & 51.30 & 38.53 \\
    & & AWQ & 67.73 & 55.47 & 28.74 & 41.37 & 59.27 &50.52\\
    & & OmniQuant & 64.79 & 51.13 & 24.83 & 40.30 & 56.90 & 47.59\\
    & & SignRound & 72.96 & 65.99 & 32.25 & 47.35 & 61.01 & 55.92 \\
    & & \cellcolor{Blue}\textbf{TesseraQ*} & \cellcolor{Blue}\textbf{75.13} & \cellcolor{Blue}\textbf{70.03} & 
    \cellcolor{Blue}\textbf{35.83} & \cellcolor{Blue}\textbf{50.17} & 
    \cellcolor{Blue}\textbf{65.19} & 
    \cellcolor{Blue}\textbf{59.27} \\
    \cmidrule{2-9}
    & \multirow{5}{*}{\shortstack{W3A16\\g128}} & GPTQ & 76.65 & 73.69 & 40.52 & 54.43 & 66.61 & 52.39 \\
    & & AWQ & 76.71 & 73.56 & 41.63 & 54.79 & 67.64 & 62.87\\
    & & OmniQuant & 76.93 & 74.66 & 39.59 & 54.95 & 67.16 & 62.66 \\
    & & SignRound & 76.82 & \textbf{75.25} & \textbf{42.92} & 55.33 & 68.27& \textbf{63.72} \\
    & & \cellcolor{Blue}\textbf{TesseraQ*} & \cellcolor{Blue}\textbf{77.58} & \cellcolor{Blue}{74.45} & \cellcolor{Blue}{41.46} & \cellcolor{Blue}\textbf{55.47} & \cellcolor{Blue}\textbf{68.90} & \cellcolor{Blue}63.59 \\
    \midrule
    \multirow{7}{*}{\shortstack{LLaMA-3.1-8B}} &  FP16 & - & 80.08 & 81.43 & 51.19 & 59.95 & 73.55 & 69.25 \\
    \cmidrule{2-9}
    & \multirow{3}{*}{\shortstack{W2A16\\g128}} & GPTQ & 53.86 & 26.55 & 20.64 & 27.87 & 53.35 & 36.46 \\
    & & AWQ & 57.34 & 35.18 & 18.26 & 28.05 & 53.27 & 38.42 \\
    & & \cellcolor{Blue}\textbf{TesseraQ*} & \cellcolor{Blue}\textbf{75.68} & \cellcolor{Blue}\textbf{68.98} & \cellcolor{Blue}\textbf{35.66} & \cellcolor{Blue}\textbf{50.21} & \cellcolor{Blue}\textbf{66.29} & \cellcolor{Blue}\textbf{59.37} \\
    \cmidrule{2-9}
    & \multirow{3}{*}{\shortstack{W3A16\\g128}} & GPTQ & 77.53 & 75.04 & 43.60 & 56.15 & 71.66 & 64.80 \\
    & & AWQ & 77.91 & 77.77 & 44.62 & 54.89 & 70.56 & 65.15 \\
    & & \cellcolor{Blue}\textbf{TesseraQ*} & \cellcolor{Blue}\textbf{79.27} &  \cellcolor{Blue}\textbf{79.46} & \cellcolor{Blue}\textbf{47.35} & \cellcolor{Blue}\textbf{57.80} & \cellcolor{Blue}\textbf{72.93} & \cellcolor{Blue}\textbf{67.36} \\
    \midrule
    \multirow{7}{*}{\shortstack{LLaMA-3.1-70B}} &  FP16 & - & 83.13 & 87.12 & 60.92 & 66.47 & 79.56 & 75.44 \\
    \cmidrule{2-9}
    & \multirow{3}{*}{\shortstack{W2A16\\g128}} & GPTQ & 65.83 & 49.54 & 26.19 & 42.74 & 61.33 & 49.11 \\
    & & AWQ & 73.45 & 68.01 & 40.27 & 48.11 & 62.19 & 58.40 \\
    & & \cellcolor{Blue}\textbf{TesseraQ*} &  \cellcolor{Blue}\textbf{78.23} &  \cellcolor{Blue}\textbf{78.70} &  \cellcolor{Blue}\textbf{47.35} &  \cellcolor{Blue}\textbf{57.91} &  \cellcolor{Blue}\textbf{71.74} &  \cellcolor{Blue}\textbf{66.79} \\
    \cmidrule{2-9}
    & \multirow{3}{*}{\shortstack{W3A16\\g128}} & GPTQ & 80.79 & 82.70 & 55.54 & 63.18 & 77.03 & 71.85 \\
    & & AWQ & 81.72 & 84.89 & 55.98 & 63.71 & 78.68 & 72.99 \\
    & & \cellcolor{Blue}\textbf{TesseraQ*} & \cellcolor{Blue}\textbf{82.86} & \cellcolor{Blue}\textbf{85.52} & \cellcolor{Blue}\textbf{58.70} & \cellcolor{Blue}\textbf{64.99} & \cellcolor{Blue}\textbf{78.37} & \cellcolor{Blue}\textbf{74.09} \\
    \bottomrule
    \end{tabular}
    \end{adjustbox}
    \label{tab_:llama_weight_only_acc}
    \vspace{-2em}
\end{table}

\textbf{Downstream Tasks Evaluation. }
We also test the weight-only quantization performance on five reasoning tasks. The results are summarized in \autoref{tab_:llama_weight_only_acc}, for LLaMA-2-7B, LLaMA-3.1-8B/70B\footnote{We did not implement OmniQaunt on LLaMA-3.1 models due to its high resource \& time demands.}. 
Notably, we found that the LLaMA-3.1-8B model demonstrates low quantization resiliency, as also shown in \cite{huang2024empiricalstudyllama3quantization}. 
For example, with W2A16g128 AWQ, this model drops more than 30\% average accuracy on downstream tasks, while the gap is 15\% for the LLaMA-2-7B model. Fortunately, our \gptbrecq can substantially increase the average performance on the downstream tasks, bringing the gap between W2 and FP16 to only 9\%. 
\gptbrecq also outperforms a recent rounding optimization method, SignRound~\citep{cheng2023optimize}, for W2A16g128, demonstrating the effectiveness of our method. 

\textbf{Weight-Activation Quantization Evaluation. }
Finally, we test weight-activation quantization scenarios with per-channel weight quantization and per-token activation quantization.  
With quantized activations, the inference speed of LLMs on GPUs/TPUs can be improved especially in the prefill stage~\citep{awq}. 
We experiment with W4A4, and W4A8 quantization and compare with three baselines, SmoothQuant, OS+, and AWQ. 
The results are provided in \autoref{tab_:llama_weight_act} (W4A8 results are provided in Appendix: \autoref{tab_:llama_weight_act_w4a8}). 
\autoref{tab_:llama_weight_act} summarizes the perplexity on WikiTex2, C4 and average accuracy on downstream tasks. (The detailed accuracy of each downstream task is located in Appendix: \autoref{tab_:llama_weight_act_acc}.)
We observe a consistent improvement of 12\% accuracy with \gptbrecq compared to AWQ. 
Additionally, we also combine our method with a recent rotation-based quantization method, QuaRot~\citep{ashkboos2024quarot}, and compare QuaRot+GPTQ and QuaRot+TesseraQ with W4A4 and W3A3 quantization. 
Combined with QuaRot, \gptbrecq also exceeds GPTQ by 10\% accuracy on 8B/70B models with W3A3 quantization, demonstrating the superiority of TesseraQ.

\begin{table}[t]
    \setlength{\tabcolsep}{8pt}
    \small
    \centering
    \caption{\textbf{W4A4/W3A3 quantization results of LLaMA-3.1}. We use per-channel weight quantization and per-token activation quantization *, $^\dagger$ means initialized from AWQ, QuaRot.}
    \begin{adjustbox}{max width=\linewidth}
    \begin{tabular}{llcccccc}
    \toprule
    \multirow{2}{*}{\textbf{Bitwidths}} & \multirow{2}{*}{\textbf{Methods}} & \multicolumn{3}{c}{\textbf{LLaMA-3.1-8B}} & \multicolumn{3}{c}{\textbf{LLaMA-3.1-70B}}\\ 
    \cmidrule(l{2pt}r{2pt}){3-5} \cmidrule(l{2pt}r{2pt}){6-8}
    & & WT2$(\downarrow)$ & C4$(\downarrow)$ & Avg. $(\uparrow)$ & WT2$(\downarrow)$ & C4$(\downarrow)$ & Avg. $(\uparrow)$\\
    \midrule
    FP16 & Pretrained & 6.24 & 9.54 & 69.25 & 2.81 & 7.11 & 75.44 \\
    \midrule
    \multirow{7}{*}{\shortstack{W4A4}} & SmoothQuant & 654.6 & 508.5 & 36.09 & 354.2  &  471.9 & 41.06 \\
    & OS+ & 124.2 & 67.44 & 40.71 & 19178 & 13750 & 35.12\\
    & AWQ & 60.99 & 74.08 & 42.51 & 24.30 & 30.39 & 53.65 \\
    & \cellcolor{Blue}\textbf{TesseraQ*} & \cellcolor{Blue}\textbf{25.73}  & \cellcolor{Blue}\textbf{30.71}  & 
    \cellcolor{Blue}\textbf{50.87} & \cellcolor{Blue}\textbf{10.45} & 
    \cellcolor{Blue}\textbf{12.77} & \cellcolor{Blue}\textbf{60.48}\\
    \cmidrule{2-8}
     & QuaRot & 17.83 & 28.08 & 51.83 & 98.01 & 149.8  & 38.92 \\
    & GPTQ$^\dagger$ & 8.39 & 13.24 & 62.87 & 5.79 & 10.12 & 68.87\\
     & \cellcolor{Blue}\textbf{{TesseraQ$^\dagger$}} & \cellcolor{Blue}\textbf{8.05} & \cellcolor{Blue}\textbf{12.62} & \cellcolor{Blue}\textbf{65.12} & 
     \cellcolor{Blue}\textbf{5.77} & \cellcolor{Blue}\textbf{10.05} & \cellcolor{Blue}\textbf{69.62}\\
    \midrule
     \multirow{3}{*}{\shortstack{W3A3}} & QuaRot & 91551 & 65662 & 35.25 & 24658 & 19155 & 34.26 \\
     & GPTQ$^\dagger$ & 93.08 & 104.73 & 37.87 & 112.3  & 162.3 & 38.02 \\
     & \cellcolor{Blue}\textbf{{TesseraQ$^\dagger$}} & \cellcolor{Blue}\textbf{27.80} & \cellcolor{Blue}\textbf{30.81} &  \cellcolor{Blue}\textbf{47.33} & \cellcolor{Blue}\textbf{38.45} & \cellcolor{Blue}\textbf{53.14} & \cellcolor{Blue}\textbf{57.42} \\
    \bottomrule
    \end{tabular}
    \end{adjustbox}
    \vspace{-2em}
    \label{tab_:llama_weight_act}
\end{table}

\textbf{Results on Mistral-7B. }
Additionally, we also test the performance of our method on the Mistral-7B model~\cite{jiang2023mistral}, which achieves high pretrained accuracy and demonstrates higher quantization resiliency. 
We test its weight-only quantization (W2A16g128, W3A16g128) and weight-activation quantization (W4A4, W4A8) performance in the Appendix (\autoref{tab_:mistral_weight_act_w4a8}). Our TesseraQ consistently outperforms other methods like SignRound, AWQ, and GPTQ.

\textbf{Results on Smaller-Size LLM for Edge Inference. }
In addition to LLMs that are deployed on GPUs, we also test the performance of smaller-size LLMs geared for edge devices. We test LLaMA-3.2-1/3B models and compare them with AWQ in \autoref{tab_:llama_3_2}. We observe that our method significantly outperforms AWQ across different bitwidths in WikiText2 perplexity and average downstream task performance.  

\begin{table}[t]
    \setlength{\tabcolsep}{8pt}
    \small
    \centering
    \caption{\textbf{Quantization Results of LLaMA-3.2 for edge inference}.}
    \begin{adjustbox}{max width=\linewidth}
    \begin{tabular}{llcccccc}
    \toprule
    \multirow{2}{*}{\textbf{Bitwidths}} & \multirow{2}{*}{\textbf{Methods}} & \multicolumn{2}{c}{\textbf{LLaMA-3.2-1B}} & \multicolumn{2}{c}{\textbf{LLaMA-3.2-3B}}\\ 
    \cmidrule(l{2pt}r{2pt}){3-4} \cmidrule(l{2pt}r{2pt}){5-6}
    & & WT2$(\downarrow)$ & Avg. $(\uparrow)$ & WT2$(\downarrow)$  & Avg. $(\uparrow)$\\
    \midrule
    FP16 & Pretrained & 9.75 & 56.50 & 7.81 & 63.57 \\
    \midrule
    \multirow{2}{*}{\shortstack{W2A16g128}} & AWQ & 5475 & 35.42 & 495.2 & 38.15  \\
    & \cellcolor{Blue}\textbf{TesseraQ*} & \cellcolor{Blue}\textbf{18.61}  & \cellcolor{Blue}\textbf{43.36}  & 
    \cellcolor{Blue}\textbf{11.94} & \cellcolor{Blue}\textbf{51.53} \\
    \cmidrule{2-6}
    \multirow{2}{*}{\shortstack{W3A16g128}} & AWQ & 16.69 & 49.85 & 10.21 & 59.94 \\
    & \cellcolor{Blue}\textbf{TesseraQ*} & \cellcolor{Blue}\textbf{11.08}  & \cellcolor{Blue}\textbf{53.24}  & 
    \cellcolor{Blue}\textbf{8.45} & \cellcolor{Blue}\textbf{61.58} \\
    \cmidrule{2-6}
    \multirow{2}{*}{\shortstack{W4A16g128}} & AWQ & 10.85 & 54.68 & 8.25 & 62.83  \\
    & \cellcolor{Blue}\textbf{TesseraQ*} & \cellcolor{Blue}\textbf{10.09}  & \cellcolor{Blue}\textbf{54.98}  & 
    \cellcolor{Blue}\textbf{7.96} & \cellcolor{Blue}\textbf{63.63} \\
    \cmidrule{2-6}
    \bottomrule
    \end{tabular}
    \end{adjustbox}
    \vspace{-1em}
    \label{tab_:llama_3_2}
\end{table}

\subsection{Ablation Studies}

\label{sec_ablation}
Below ablation studies are conducted with the LLaMA-2-7B model with W2A16g128 quantization. 

\textbf{Calibration Data. }
In this section, we compare the performance of different calibration datasets and sizes. We sample calibration data from either WikiText2~\citep{wikitext2} or C4~\citep{c4} training dataset. 
We also experiment with the different sample sizes, ranging from 128 to 512. Meanwhile, we change the batch size during rounding optimization, ranging from 1 to 4. 

\autoref{tab_:ablation_studies} demonstrates the task performance (PPL and average accuracy metric) as well as the calibration costs (algorithm runtime and GPU memory footprint). 
First, we find that the source of calibration data will impact the perplexity evaluation. The performance benefits if evaluation data and calibration data are from the same dataset. For example, the C4-calibrated model has 1.2 higher WikiText2 PPL than the WikiText2-calibrated model. 
Second, increasing the number of samples and the batch size consistently improves the task performance. However, it may also lead to higher runtime and GPU memory consumption, which may be alleviated via multi-GPU calibration. Nevertheless, it is worthwhile to note that even with 128 samples and a batch size of 1, our \gptbrecq can significantly improve the baseline AWQ results. 

\begin{table}[t]
    \setlength{\tabcolsep}{7pt}
    \small
    \centering
    \caption{\textbf{Ablation studies of calibration data source and data sizes}. We report the LLaMA-2-7B W2A16g128 quantization results with task performances and calibration costs.}
    \begin{adjustbox}{max width=\linewidth}
    \begin{tabular}{llcccccccc}
        \toprule
        \multirow{2}{*}{\textbf{\#Samples}} & \multirow{2}{*}{\textbf{BS}} & \textbf{Runtime/} & \multicolumn{3}{c}{\textbf{Calib. Data: WikiText2}} & \multicolumn{3}{c}{\textbf{ Calib. Data: C4}}\\ 
        \cmidrule(l{2pt}r{2pt}){4-6} \cmidrule(l{2pt}r{2pt}){7-9}
        & & \textbf{GPU Mem.} & WikiText2$(\downarrow)$ & C4$(\downarrow)$ & Avg.$(\uparrow)$ & WikiText2$(\downarrow)$ & C4$(\downarrow)$ & Avg.$(\uparrow)$\\
        \midrule
        128 & 1 & 3.2h/17.5GB & 7.33 & 11.39 & 56.58 & 8.54 & 10.83 & 56.87 \\
        256 & 2 & 3.9h/28.6GB &  7.10 & 11.16 & 57.17 &8.32 & 10.66 & 57.85 \\
        512 & 2 & 4.0h/40.4GB & 7.14 & 11.22 & 57.42 &  8.22 & 10.47 & 58.56\\
        512 & 4 & 6.0h/65.4GB & \textbf{6.82} & 10.77 & 58.35 & 8.05 & \textbf{10.29} & \textbf{59.27} \\
        \bottomrule
    \end{tabular}
    \end{adjustbox}
    \label{tab_:ablation_studies}
\end{table}


\newcommand{\cmark}{\ding{51}}%
\newcommand{\xmark}{\ding{55}}%

\textbf{Algorithm choices. }
We also test the algorithm choices in \gptbrecq. To be more specific, we experiment with block reconstruction with or without progressive adaptive rounding (PAR) and dequantization scale tuning (DST) and compare their final task performance.
As shown in \autoref{wrap-tab:1}, both PAR and DST contribute a lot to the final perplexity metric (denoted \begin{wraptable}{r}{5.5cm}
\vspace{-1.5em}
\caption{TesseraQ Algorithm choices.}\label{wrap-tab:1}
\vspace{-1em}
\begin{adjustbox}{max width=\linewidth}
\begin{tabular}{llrrr}\\
\toprule 
\textbf{PAR} & \textbf{DST} & \textbf{WT2} & \textbf{C4} & \textbf{Avg.} \\
\midrule
\xmark & \xmark & 14.65 & 18.67 & 50.52\\  
\cmark & \xmark & 7.72 & 11.95 & 56.79 \\
\xmark & \cmark & 8.58 & 13.14 & 54.45\\  
\cmark & \cmark & \textbf{6.82} & \textbf{10.77} & \textbf{58.35} \\
\bottomrule
\end{tabular}  
\end{adjustbox}
\vspace{-1.4em}
\end{wraptable} by WT2 (WikiText2) and C4) and average accuracy (denoted by Avg.). Remarkably, applying one of them solely can also improve the AWQ baseline (first row) results by a large margin. 

\begin{figure}[t]
\centering
\includegraphics[width=\linewidth]{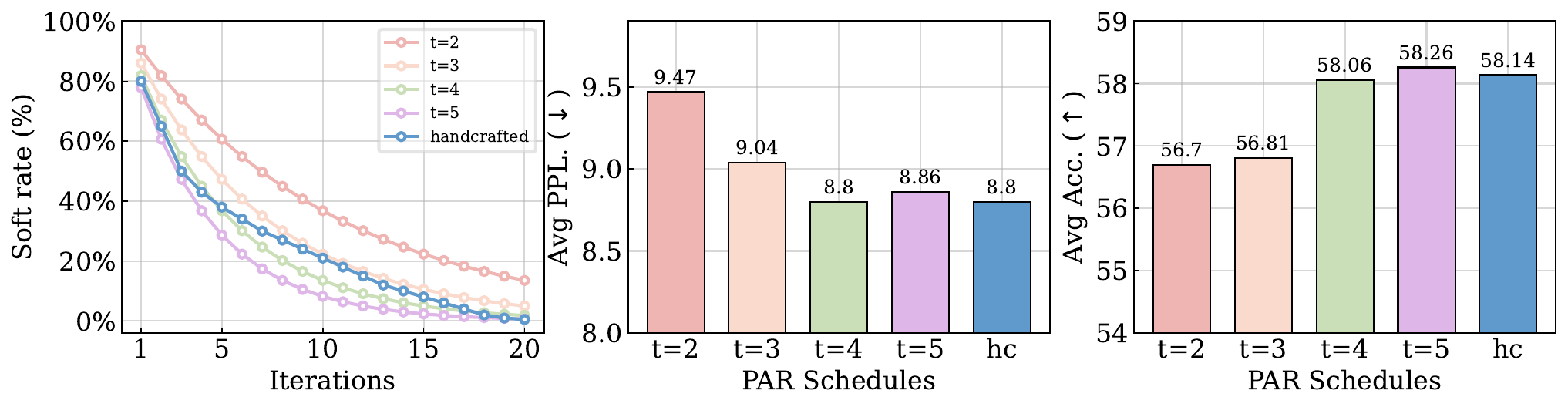}
\vspace{-1em}
\caption{\textbf{Ablation study of PAR schedule.} We experiment several rule-based $P$ adjustments and one handcrafted adjustment. \emph{(AWQ baseline results: average PPL: 16.66, average acc.: 50.52).} } 
\vspace{-1em}
\label{fig_par_schedule}

\end{figure}
\textbf{PAR Schedule. }
We investigate how to adjust the $P$ during progressive adaptive rounding. In our implementation, we use a handcrafted design, which manually decreases the soft rate (i.e., the percentage of soft rounding variable) as shown in \autoref{fig_par_schedule}. Our handcrafted design gradually decays the soft rate. To demonstrate that our PAR is quite robust to the schedule of soft rate, we also test several rule-based adjustments, which adjust the soft rate as $\frac{1}{\mathrm{exp}(tx)}$, where $x\in(0, 1]$ is the scaled iteration number and $t$ is the temperature hyper-parameter. We test $t=\{2, 3, 4, 5\}$ and compare it with our handcrafted implementation with LLaMA-2-7B W2A16g128 quantization. The results in \autoref{fig_par_schedule} show that $t=4, 5$ and our handcrafted adjustments obtain the best performance. Overall, we find that our algorithm is not sensitive to the scheduling, and has consistently superior performance than the AWQ initialized model.

\subsection{Visualization}
\begin{figure}[t]
\centering
\includegraphics[width=\linewidth]{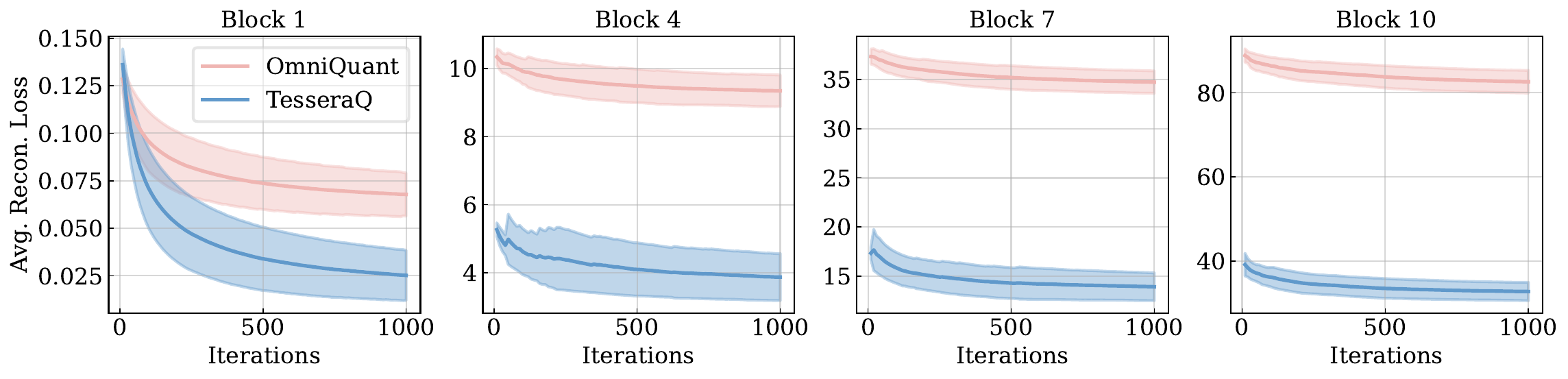}
\vspace{-2em}
\caption{\textbf{Reconstruction loss convergence.} We compare the block reconstruction loss of OmniQuant and TesseraQ during optimization. Our method significantly reduces the loss in each block.  } 
\label{fig_converge}
\end{figure}

In this section, we provide visualizations of our calibration process to interpret the effectiveness of our method. The experiments are conducted on LLaMA-2-7B with W2A16g128 quantization. 
We first compare the loss convergence value in OmniQuant and TesseraQ, both of which calibrate the model with block reconstruction loss. 
To ensure a fair comparison, we use the same AWQ initialization to these two methods and align all training hyper-parameters. 
As shown in \autoref{fig_converge}, during the first block reconstruction, \gptbrecq reduces more loss than OmniQuant. In the following blocks, the loss gap between our method and Omniquant keeps on increasing. 
Consequently, \gptbrecq will have a much lower model output error due to the cumulative effect of reconstruction.

Since rounding variables ($\alpha$) are binary, we also demonstrate the number or percentage of rounding variables that flip after TesseraQ. In \autoref{tab_rounding}, we show the number and the percentage of flipped variables. Overall we observe around 3\%$\sim$8\% of variables flip, amounting to over 10M parameters per block. This is a significantly larger value than AWQ/OmniQuant, which justifies why TesseraQ further improves the performance. 
We also found that attention layers tend to have less flipped rounding compared to MLP layers. 2/3-bit quantization also flips more than 4-bit quantization.

\vspace{-0.5em}
\subsection{Hardware Evaluation}
\vspace{-0.5em}

To demonstrate the weight compression effect and the inference throughput change, we test LLaMA-3.1-8B/70B/405B under different GPU environments, kernel backend and different bitwidths. \autoref{tab_hardware} summarizes the results of inference throughput (generated token per second) with batch size 1 or 16. Remarkably, W2A16g128 reduces the weight memory of the 405B model from 756GB to 114 GB and the 70B model from 132 GB to 21 GB. However, the INT2 dequantization kernel (in Triton~\citep{Triton} support) is currently less optimized, especially for larger models, expending lower throughput compared to FP16. We find that INT4 with Exllama kernel can increase the throughput when batch size is 1 and achieve similar throughput with FP16 model when batch size is 16. Nonetheless, it is worthwhile to note that our TesseraQ complies with standard uniform quantization formats and can be deployed with various kernels that support uniform quantization on various devices, e.g., GPU, CPU, TPU, edge processor.

\begin{table}[t]
\setlength{\tabcolsep}{3pt}
\small
\centering
\caption{\textbf{Number (percentages) of rounding variables that flip after TesseraQ}. We average the numbers across all blocks on LLaMA-2-7B.}
\begin{adjustbox}{max width=\linewidth}
\begin{tabular}{lccccccc}
    \toprule
    \textbf{Bits/Layers} & \textbf{q\_proj} & \textbf{k\_proj} & \textbf{v\_proj}  & \textbf{o\_proj}  & \textbf{gate\_proj} & \textbf{up\_proj} & \textbf{down\_proj}\\ 
    \midrule
    W4A16g128 & 498k (2.97\%) & 477k (2.85\%) & 520k (3.10\%) & 620k (3.70\%) &  1.77M (3.92\%) & 1.81M (4.02\%) & 1.91M (4.24\%) \\
    W2A16g128 & 765k (4.55\%) & 734k (4.37\%) & 758k (4.52\%) & 961k (5.73\%) & 3.00M (6.67\%) & 2.99M (6.64\%) & 3.21M (7.12\%) \\
    \bottomrule
\end{tabular}
\end{adjustbox}
\vspace{-2em}
\label{tab_rounding}
\end{table}

\begin{table}[t]
\setlength{\tabcolsep}{7pt}
\small
\centering
\caption{\textbf{Comparison of weight memory compression and inference throughput}. We measure LLaMA-3.1 series model under various bitwidth/backend. WM stands for weight memory, $TP_{n}$ denotes inference throughput with a batch size of $n$ (output token/s). }
\begin{adjustbox}{max width=\linewidth}
\begin{tabular}{llccccccccc}
    \toprule
    \textbf{LLaMA-3.1} &  & \multicolumn{3}{c}{\textbf{8B (1$\times$A5000)}} & \multicolumn{3}{c}{\textbf{70B (2$\times$A100-80GB)}}  & \multicolumn{3}{c}{\textbf{405B (4$\times$A100-80GB)}} \\ 
    \cmidrule(l{2pt}r{2pt}){3-5} \cmidrule(l{2pt}r{2pt}){6-8} \cmidrule(l{2pt}r{2pt}){9-11}
    BitWidth & Backend &  WM & $TP_1$ & $TP_{16}$ & WM & $TP_1$ & $TP_{16}$ & WM &$TP_1$ & $TP_{16}$\\
    \midrule
    FP16 & Pytorch & 15GB & 49.23 & 358.1 & 132GB & 12.31 & 104.0 & 756GB & OOM & OOM \\
    \midrule 
    W4A16g128 & Exllama & 5.5GB & 57.54 & 361.1 & 39GB & 26.23 & 86.94 & 209GB & 7.01 & 18.59 \\
    W2A16g128 & Triton & 3.9GB & 165.3 & 545.5 & 21GB & 4.93 &  54.35 & 114GB & 0.18 & 2.94 \\
    \bottomrule
\end{tabular}
\end{adjustbox}
\vspace{-2em}
\label{tab_hardware}
\end{table}

\vspace{-0.5em}
\section{Related Work}
\vspace{-0.5em}

Quantization has been a primary method to compress and accelerate off-the-shelf large models. 
Survey papers by \cite{gholami2022survey} and \cite{nn2} have systematically summarized the progress of quantization. 
Here, we list several major quantization works, especially for LLMs. 

\textbf{Post-Training Quantization for LLMs. }
While Quantization aware Training (QAT) guarantees better task performance in low-bit quantization, PTQ is more suitable for LLM due to its less reliance on computing resources and training data. 
PTQ methods like \cite{gptq, awq, outlier, outlier-plus, shao2023omniquant, quip, qllm} improve the uniform quantization performance by optimizing weights, transformation scales, and clipping ranges. Our method continues improving the uniform quantization effect by incorporating rounding optimization. 
Other works try to improve PTQ in LLMs in different ways. For example, AQLM and GPTVQ~\citep{egiazarian2024extreme, van2024gptvq} explore non-uniform quantization schemes for weight-only quantization, which may better match the distribution of weights. 
LLM.int8~\citep{llmint8}, BiLLM~\citep{huang2024billm}, SiLLM~\citep{huang2024slim} apply mixed-precision quantization to keep salient weights in high precision and maintain the accuracy. However, these methods cannot be applied to quantize activations and thus cannot support integer MatMul. 
QuaRot~\citep{ashkboos2024quarot}, SpinQuant~\citep{liu2024spinquant} target activation outliers and eliminate them through the rotation matrix. We have demonstrated that our method can also be combined with them.

\textbf{QAT for LLM. }
Recent works also explore QAT-based quantization for LLMs. To reduce data access, LLM-QAT~\citep{llmqat} generates language data for data-free QAT. 
To prevent massive weight memory usage, Q-LoRA~\citep{qlora} applies quantization-aware low-rank adaptation for finetuning. 
Recently, BitNet and BitNet b.158~\citep{wang2023bitnet, ma2024era} trained a 1-bit and 1.58-bit model from scratch, enabling multiplication-free LLM. However, these methods are hard to scale up due to the massive memory and computation requirements, especially for more than 70B models. As a result, they only focus on 1B$\sim$3B-scale models.

\section{Conclusion}

In this paper, we have proposed TesseraQ, a PTQ method for effectively calibrating large language models. Based on block reconstruction, TesseraQ optimizes weight rounding through a progressive approach that iteratively hardens and softens the rounding variables. 
Together with dequantization scale tuning, TesseraQ can be seamlessly combined with other PTQ methods like transformation, clipping, and rotation, to reach new state-of-the-art performance. We demonstarte TesseraQ's superiority on open source LLaMA models. TesseraQ establishes a new state-of-the-art for quantized LLMs, in terms of perplexity, downstream accuracy and hardware performance.

\subsubsection*{Acknowledgments}
This work was supported in part by CoCoSys, a JUMP2.0 center sponsored by DARPA and SRC, the National Science Foundation (CAREER Award, Grant \#2312366, Grant \#2318152), the DARPA Young Faculty Award and the DoE MMICC center SEA-CROGS (Award \#DE-SC0023198). We would also like to thank Amir Yazdanbaksh, Joao Carreira and James Laudon for their helpful feedback and insightful suggestions towards this work.

\bibliography{iclr2025_conference}
\bibliographystyle{iclr2025_conference}

\newpage
\appendix
\section{More Experimental Results}

In this section, we include additional experimental results from the main section. 

\subsection{Results on C4}
We demonstrate the perplexity results on the C4 datasets in \autoref{tab_:llama_weight_only_c4}. Note that the OmniQuant results are re-evaluated using the official checkpoint, which is slightly higher than the original paper results~\citep{shao2023omniquant}. 
Since the evaluation protocol can be different across different papers,
we ensure use of the same evaluation protocol to compare different methods.  
Note, we restrict all models here from using the WikiText2 calibration data as the calibration data will affect the perplexity metric as shown in our ablation study.
The improvements of our method over existing approaches are consistent with the results on the WikiText2 dataset. 

\begin{table}[h]
    \setlength{\tabcolsep}{10pt}
    \small
    \centering
    \caption{\textbf{Weight-only quantization results of LLaMA-1 and LLaMA-2 Models}. We report C4 perplexity in this table. *, $\dag$ means initialized from AWQ, and OmniQuant, respectively.}
    \begin{adjustbox}{max width=\linewidth}
    \begin{tabular}{llccccccc}
        \toprule
          \multicolumn{2}{l}{\textbf{LLaMA1\&2 / PPL}$\downarrow$} & 1$-$7B & 1$-$13B & 1$-$30B & 1$-$65B & 2$-$7B & 2$-$13B &2$-$70B \\  \midrule
        FP16 & - & 7.08 & 6.61 & 5.98 & 5.62 & 6.97 & 6.46 & 5.52 \\ 
        \midrule
        \multirow{4}{*}{\shortstack{W2A16}} & RTN & 1.3e5 & 5.6e4 & 2.7e4 & 2.2e4 & 4.8e4 & 7.2e4 & 2.4e4  \\
         & GPTQ  & 689.13 & 2.5e3  & 169.80 & 40.58 & NAN  & 323.12 & 48.82 \\
         & OmniQuant & 26.03 & 18.94 & 14.55 & 11.47 & 90.64 & 26.76 & 13.33 \\ 
         & \cellcolor{Blue}\textbf{TesseraQ$^\dag$}  & \cellcolor{Blue}\textbf{13.28} & \cellcolor{Blue}\textbf{11.43} & \cellcolor{Blue}\textbf{10.81} &  \cellcolor{Blue}\textbf{8.52} & \cellcolor{Blue}\textbf{14.82} & \cellcolor{Blue}\textbf{11.96} & \cellcolor{Blue}\textbf{9.15}   \\
        \midrule
        \multirow{5}{*}{\shortstack{W2A16\\g128}} & RTN & 1.0e3 & 447.64 & 99.45 & 17.15 & 4.9e3 & 139.65 & 42.13  \\
         & GPTQ & 27.71 & 15.29 & 11.93 & 11.99 & 33.70 & 20.97 & NAN  \\
         & AWQ & 16.35  & 12.93 & 10.07 & 8.78 & 18.67 & 11.88  & 8.49  \\
         & OmniQuant & 14.06 & 11.27 & 10.37 & 8.65 & 16.34 & 12.14 & 9.33 \\
         & \cellcolor{Blue}\textbf{TesseraQ*}  & \cellcolor{Blue}\textbf{10.64} & \cellcolor{Blue}\textbf{9.36} & \cellcolor{Blue}\textbf{8.36} & \cellcolor{Blue}\textbf{7.64} & \cellcolor{Blue}\textbf{10.77} & \cellcolor{Blue}\textbf{9.48} & \cellcolor{Blue}\textbf{7.63}  \\
        \midrule
        \multirow{5}{*}{\shortstack{W2A16\\g64}} & RTN & 151.43 & 76.00 & 30.07 & 11.34 & 475.35 & 28.69 & 13.43  \\
         & GPTQ & 17.71 & 11.70 & 9.92 & 10.07 & 19.40 & 12.48 & NAN  \\
         & AWQ &  13.47 & 11.35 & 9.12 & 8.11 & 15.13 & 10.85 & 7.77 \\
         & OmniQuant & 12.79 & 10.60 & 9.46 & 8.18 & 13.79 & 11.02 & 8.61 \\
         & \cellcolor{Blue}\textbf{TesseraQ*}  & \cellcolor{Blue}\textbf{10.32}  & \cellcolor{Blue}\textbf{9.05} & \cellcolor{Blue}\textbf{8.18} & \cellcolor{Blue}\textbf{7.48}   & \cellcolor{Blue}\textbf{10.50} & \cellcolor{Blue}\textbf{9.23} & \cellcolor{Blue}\textbf{7.44} \\ \midrule
        \multirow{4}{*}{\shortstack{W3A16}} & RTN & 28.26 & 13.22 & 28.66 & 12.79 & 402.35 & 12.51 & 10.02  \\
         & GPTQ & 9.49 & 8.16 & 7.29 & 6.71 & 9.81 & 8.02 & 6.57  \\
         & AWQ & 11.16 & 8.37 & 7.91 & 8.62 & 16.25 & 8.90 & 6.50 \\
         & OmniQuant & 8.73 & 7.68 & 6.86 & 6.31 & 9.24 & 7.89 & 6.31 \\
         & \cellcolor{Blue}\textbf{TesseraQ*}  & \cellcolor{Blue}\textbf{8.15} & \cellcolor{Blue}\textbf{7.38} & \cellcolor{Blue}\textbf{6.60} & \cellcolor{Blue}\textbf{6.16}  & \cellcolor{Blue}\textbf{8.30} & \cellcolor{Blue}\textbf{7.41} & \cellcolor{Blue}\textbf{6.08} \\ 
        \bottomrule
    \end{tabular}
    \end{adjustbox}
    \label{tab_:llama_weight_only_c4}
\end{table}

\subsection{W4A8 Quantization }

We also provide the W4A8 quantization in \autoref{tab_:llama_weight_act_w4a8}. Overall we find a small difference in W4A8 quantization due to the 8-bit per-token activation quantization. 

\begin{table}[h]
    \setlength{\tabcolsep}{8pt}
    \small
    \centering
    \caption{\textbf{Weight-activation quantization Results of various LLMs}. We report the accuracy of 5 reasoning tasks $(\uparrow)$.}
    \begin{adjustbox}{max width=\linewidth}
    \begin{tabular}{lclcccccc}
    \toprule
    \textbf{Models} & \textbf{Bitwidths} & \textbf{Methods} & \textbf{PiQA} & \textbf{ArcE} & \textbf{ArcC} & \textbf{HellaSwag} & \textbf{WinoGrande} & \textbf{Avg.} \\ 
    \midrule
    \multirow{5}{*}{\shortstack{LLaMA-7B}} & FP16 & -&77.47 & 52.48 & 41.46 & 73.00 & 67.07 & 62.30 \\
    \cmidrule{2-9}
    & \multirow{4}{*}{\shortstack{W4A8}} & SmoothQuant & 75.19 & 70.45 & 37.45 & 51.06 & 64.87 & 59.81 \\
    & & OS+ & 78.42 & 74.49 & 40.61 & 55.53 & 69.37 & 63.75 \\
    & & AWQ & 77.63 & 73.31 & 41.89 & 55.50 & 69.85 & 63.65 \\
    & & \cellcolor{Blue}\textbf{TesseraQ*} & \cellcolor{Blue}\textbf{78.89} & \cellcolor{Blue}\textbf{75.33} & 
    \cellcolor{Blue}\textbf{41.55} & \cellcolor{Blue}\textbf{56.11} & 
    \cellcolor{Blue}\textbf{69.14} & 
    \cellcolor{Blue}\textbf{64.21} \\
    \midrule
    \multirow{5}{*}{\shortstack{LLaMA-2-7B}} &  FP16 & - & 78.07 & 76.34 & 43.51 & 57.17 & 69.21 & 64.87 \\
    \cmidrule{2-9}
    & \multirow{4}{*}{\shortstack{W4A8}} & SmoothQuant & 75.24 & 70.95 & 38.39 & 51.30 & 63.85 & 59.95 \\
    & & Outlier Supp.+ & 77.09 & 74.74 & 42.57 & 56.37 & 68.51 & 63.86 \\
    & & AWQ & 77.09 & 74.36 & 42.32 & 56.25 & 69.53 & 63.91 \\
    & & \cellcolor{Blue}\textbf{TesseraQ*} & \cellcolor{Blue}\textbf{77.42} & \cellcolor{Blue}{76.26} & \cellcolor{Blue}{41.63} & \cellcolor{Blue}\textbf{56.42} & \cellcolor{Blue}\textbf{69.22} & \cellcolor{Blue}\textbf{64.19} \\
    \midrule
    \multirow{5}{*}{\shortstack{LLaMA-3.1-8B}} &  FP16 & - & 79.54 & 80.09 & 50.17 & 60.13 & 73.24 & 68.64 \\
    \cmidrule{2-9}
    & \multirow{4}{*}{\shortstack{W4A8}} & SmoothQuant & 71.98 & 66.37 & 34.55 & 50.46 & 67.40 & 58.16 \\
    & & Outlier Supp.+ & 77.91 & 78.78 & 48.03 & 58.83 & 72.53 & 67.22 \\
    & & AWQ & 79.00 & 78.40 & 48.63 & 58.81 & 72.45 & 67.46 \\
    & & \cellcolor{Blue}\textbf{TesseraQ*} & \cellcolor{Blue}\textbf{78.99} & \cellcolor{Blue}{79.88} & \cellcolor{Blue}{47.61} & \cellcolor{Blue}\textbf{59.09} & \cellcolor{Blue}\textbf{72.77} & \cellcolor{Blue}\textbf{67.67} \\
    \midrule
   \multirow{4}{*}{\shortstack{Mistral-8B}} & \multirow{4}{*}{\shortstack{W4A8}} & SmoothQuant & 79.59 & 77.56 & 46.50 & 57.62 & 71.11 & 66.48 \\
    & & OS+ & 80.35 & 79.04 & 48.03 & 60.18 & 72.45 & 68.02 \\
    & & AWQ & 79.92 & 79.79 & 47.35 & 58.80 & 74.26 & 68.03 \\
    & & \cellcolor{Blue}\textbf{TesseraQ*} & \cellcolor{Blue}\textbf{80.36} & \cellcolor{Blue}\textbf{79.92} & 
    \cellcolor{Blue}\textbf{49.57} & \cellcolor{Blue}\textbf{60.54} & 
    \cellcolor{Blue}\textbf{73.79} & 
    \cellcolor{Blue}\textbf{68.84} \\
    \bottomrule
    \end{tabular}
    \end{adjustbox}
    \label{tab_:llama_weight_act_w4a8}
\end{table}

\begin{table}[t]
    \setlength{\tabcolsep}{8pt}
    \small
    \centering
    \caption{\textbf{Weight-activation quantization Results of Mistral-7B}. We report the accuracy of 5 reasoning tasks $(\uparrow)$.}
    \begin{adjustbox}{max width=\linewidth}
    \begin{tabular}{lclcccccc}
    \toprule
    \textbf{Models} & \textbf{Bitwidths} & \textbf{Methods} & \textbf{PiQA} & \textbf{ArcE} & \textbf{ArcC} & \textbf{HellaSwag} & \textbf{WinoGrande} & \textbf{Avg.} \\ 
    \midrule
    \multirow{15}{*}{\shortstack{Mistral-7B}} &  FP16 & - &  80.68 & 80.93 & 50.42 & 61.26 & 73.79 & 69.42 \\
    \cmidrule{2-9}
    & \multirow{4}{*}{\shortstack{W2A16\\g128}} & GPTQ & 64.20 & 45.74 & 22.35 & 36.68 & 55.02 & 44.80 \\
    & & AWQ & 68.44 & 56.73 & 27.44 & 40.60 & 56.03 & 49.06 \\
    & & SignRound & 75.84 & 70.88 & 30.73 & 50.87 & 62.90 & 58.24 \\
    & & \cellcolor{Blue}\textbf{TesseraQ*} & \cellcolor{Blue}\textbf{76.87} & \cellcolor{Blue}\textbf{71.67} & \cellcolor{Blue}\textbf{39.59} & \cellcolor{Blue}\textbf{54.09} & \cellcolor{Blue}\textbf{68.11} & \cellcolor{Blue}\textbf{62.07} \\
    \cmidrule{2-9}
    & \multirow{3}{*}{\shortstack{W3A16\\g128}} & GPTQ & 79.70 & 78.70 & 48.41 & 59.15 & 71.98 & 67.19 \\
    & & AWQ & 80.19 & 78.62 & 45.56 & 58.28 & 71.58 & 66.85 \\
    & & SignRound & 79.54 & 78.70 & 46.33 & 59.60 & 72.85 & 67.40 \\
    & & \cellcolor{Blue}\textbf{TesseraQ*} & \cellcolor{Blue}\textbf{79.59} &  \cellcolor{Blue}\textbf{78.36} & \cellcolor{Blue}\textbf{47.44} & \cellcolor{Blue}\textbf{59.87} & \cellcolor{Blue}\textbf{71.98} & \cellcolor{Blue}\textbf{67.45} \\
    \cmidrule{2-9}
    & \multirow{4}{*}{\shortstack{W4A4}} & SmoothQuant & 57.94 & 35.14 & 21.75 & 30.51 & 48.30 & 38.73 \\
    & & OS+ & 66.70 & 56.73 & 30.20 & 42.39 & 52.01 & 49.61 \\
    & & AWQ & 66.26 & 54.16 & 30.80 & 43.45 & 53.67 & 49.67 \\
    & & \cellcolor{Blue}\textbf{{TesseraQ*}} & \cellcolor{Blue}\textbf{72.19} & \cellcolor{Blue}\textbf{65.90} & 
    \cellcolor{Blue}\textbf{33.78} & \cellcolor{Blue}\textbf{49.02} & 
    \cellcolor{Blue}\textbf{57.61} & 
    \cellcolor{Blue}\textbf{55.71} \\
    \bottomrule
    \end{tabular}
    \end{adjustbox}
    \label{tab_:mistral_weight_act_w4a8}
\end{table}

\subsection{Detailed Accuracy of W4A4/W3A3 Quantization}

\autoref{tab_:llama_weight_act_acc} provides the detailed accuracy of each zero-shot tasks in W4A4/W3A3 quantization. 

\begin{table}[h]
    \setlength{\tabcolsep}{8pt}
    \small
    \centering
    \caption{\textbf{W4A4/W3A3 quantization results of LLaMA-3.1}. We use per-channel weight quantization and per-token activation quantization *, $^\dagger$ means initialized from AWQ, QuaRot.}
    \begin{adjustbox}{max width=\linewidth}
    \begin{tabular}{lclcccccc}
    \toprule
    \textbf{Models} & \textbf{Bitwidths} & \textbf{Methods} & \textbf{PiQA} & \textbf{ArcE} & \textbf{ArcC} & \textbf{HellaSwag} & \textbf{WinoGrande} & \textbf{Avg.} \\ 
    \midrule
    \multirow{11}{*}{\shortstack{LLaMA-3.1-8B}} &  FP16 & - & 79.54 & 80.09 & 50.17 & 60.13 & 73.24 & 68.64 \\
    \cmidrule{2-9}
    & \multirow{7}{*}{\shortstack{W4A4}} & SmoothQuant & 54.24 & 27.90 & 19.79 & 26.87 & 51.61 & 36.09 \\
    & & OS+ & 57.34 & 40.99 & 20.22 & 33.19 & 51.77 & 40.71  \\
    & & AWQ & 59.68 & 44.90 & 22.09 & 34.53 & 51.30 & 42.51 \\
    & & \cellcolor{Blue}\textbf{TesseraQ*} & \cellcolor{Blue}\textbf{67.08} & \cellcolor{Blue}\textbf{59.09} & 
    \cellcolor{Blue}\textbf{27.13} & \cellcolor{Blue}\textbf{43.88} & 
    \cellcolor{Blue}\textbf{57.14} & 
    \cellcolor{Blue}\textbf{50.87} \\
    & & QuaRot & 69.85 & 58.03 & 28.07 & 43.37 & 59.82 & 51.83 \\
    & & GPTQ$^\dagger$ & 76.22 & 73.94 & 41.21 & 55.47 & 67.48 & 62.87 \\
    & & \cellcolor{Blue}\textbf{{TesseraQ$^\dagger$}} & \cellcolor{Blue}\textbf{77.64} & \cellcolor{Blue}\textbf{77.27} & \cellcolor{Blue}\textbf{44.80} & \cellcolor{Blue}\textbf{56.03} & \cellcolor{Blue}\textbf{69.85} & \cellcolor{Blue}\textbf{65.12} \\
    \cmidrule{2-9}
    & \multirow{3}{*}{\shortstack{W3A3}} & QuaRot & 52.28 & 26.59 & 20.56 & 26.11 & 50.67 & 35.25 \\
    & & GPTQ$^\dagger$ & 56.96 & 33.62 & 20.47 & 28.87 & 49.40 & 37.87 \\
    & & \cellcolor{Blue}\textbf{{TesseraQ$^\dagger$}} & \cellcolor{Blue}\textbf{66.05} & \cellcolor{Blue}\textbf{51.59} & 
    \cellcolor{Blue}\textbf{24.40} & \cellcolor{Blue}\textbf{40.59} & 
    \cellcolor{Blue}\textbf{53.98} & 
    \cellcolor{Blue}\textbf{47.33} \\
    \midrule
    \multirow{10}{*}{\shortstack{LLaMA-3.1-70B}} &  FP16 & - & 83.13 & 87.12 & 60.92 & 66.47 & 79.56 & 75.44 \\
    \cmidrule{2-9}
    & \multirow{7}{*}{\shortstack{W4A4}} & SmoothQuant & 57.45 & 38.46 & 24.23 & 30.22 & 54.93 & 41.06 \\
    & & OS+ & 53.04 & 25.79 & 22.01 & 25.88 & 48.85 & 35.12 \\
    & & AWQ & 69.91 & 61.71 & 34.04 & 47.98 & 54.61 & 53.65 \\
    & & \cellcolor{Blue}\textbf{{TesseraQ*}} & \cellcolor{Blue}\textbf{78.29} & \cellcolor{Blue}\textbf{69.15} & \cellcolor{Blue}\textbf{38.12} & \cellcolor{Blue}\textbf{53.74} & 
    \cellcolor{Blue}\textbf{61.16} & 
    \cellcolor{Blue}\textbf{60.09} \\
     & & QuaRot & 57.88 & 36.36 & 19.02 & 28.13 & 53.19 & 38.92\\
     & & GPTQ$^\dagger$ & 79.76 & 80.17 & 50.59 & 60.71 & 73.08 & 68.87 \\
     & & \cellcolor{Blue}\textbf{{TesseraQ$^\dagger$}} & \cellcolor{Blue}\textbf{81.84} & \cellcolor{Blue}\textbf{82.64} & 
     \cellcolor{Blue}\textbf{54.07} & \cellcolor{Blue}\textbf{63.90} & 
     \cellcolor{Blue}\textbf{65.64} & 
     \cellcolor{Blue}\textbf{69.62} \\
     \cmidrule{2-9}
    & \multirow{3}{*}{\shortstack{W3A3}} & QuaRot & 52.06 & 24.87 & 20.05 & 25.55 & 49.25 & 34.26 \\
    & & GPTQ$^\dagger$ & 55.98 & 34.80 & 19.45 & 28.38 & 51.46 & 38.02  \\
    & & \cellcolor{Blue}\textbf{{TesseraQ$^\dagger$}} & \cellcolor{Blue}\textbf{74.80} & \cellcolor{Blue}\textbf{66.03} & 
    \cellcolor{Blue}\textbf{36.42} & \cellcolor{Blue}\textbf{51.34} &  \cellcolor{Blue}\textbf{58.43} &
    \cellcolor{Blue}\textbf{57.42} \\
    \bottomrule
    \end{tabular}
    \end{adjustbox}
    \vspace{-2em}
    \label{tab_:llama_weight_act_acc}
\end{table}
\end{document}